\documentclass[preprint,12pt]{elsarticle}

\usepackage{amssymb}
\usepackage{subfigure}
\usepackage{hyperref}
\usepackage{bm}
\usepackage{stfloats}
\usepackage{multirow}
\usepackage[ruled,linesnumbered]{algorithm2e}
\usepackage{booktabs}

\journal{Knowledge-Based Systems}

\begin{document}

\begin{frontmatter}



\title{Semantically Consistent Multi-view Representation Learning}


\author[label1]{Yiyang Zhou}
\author[label2]{Qinghai Zheng}
\author[label1]{Shunshun Bai}
\author[label1]{Jihua Zhu\corref{cor1}}
\cortext[cor1]{Corresponding author, email: zhujh@xjtu.edu.cn.}
\affiliation[label1]{organization={School of Software Engineering},
            addressline={Xi’an Jiaotong University}, 
            city={Xi’an},
            postcode={710049}, 
            country={China}}
            
\affiliation[label2]{organization={College of Computer and Data Science},
            addressline={Fuzhou University}, 
            city={Fuzhou},
            postcode={350108}, 
            country={China}}

\begin{abstract}
In this work, we devote ourselves to the challenging task of Unsupervised Multi-view Representation Learning (UMRL), which requires learning a unified feature representation from multiple views in an unsupervised manner. Existing UMRL methods mainly concentrate on the learning process in the feature space while ignoring the valuable semantic information hidden in different views. To address this issue, we propose a novel Semantically Consistent Multi-view Representation Learning (SCMRL), which makes efforts to excavate underlying multi-view semantic consensus information and utilize the information to guide the unified feature representation learning. Specifically, SCMRL consists of a within-view reconstruction module and a unified feature representation learning module, which are elegantly integrated by the contrastive learning strategy to simultaneously align semantic labels of both view-specific feature representations and the learned unified feature representation. In this way, the consensus information in the semantic space can be effectively exploited to constrain the learning process of unified feature representation. Compared with several state-of-the-art algorithms, extensive experiments demonstrate its superiority. Our code is released on \href{https://github.com/YiyangZhou/SCMRL}{https://github.com/YiyangZhou/SCMRL}.

\end{abstract}



\begin{keyword}
Multi-view representation learning \sep Contrastive learning \sep Semantic consensus information


\end{keyword}

\end{frontmatter}


\section{Introduction}
\label{}
Multi-view data are prevalent in real-world applications, and different views can be collected from diverse sensors or various feature extractors. However, numerous classic and effective algorithms \cite{shi2000normalized, macqueen1967classification, xie2016unsupervised} are designed for single-view data and can not be leveraged to multi-view data directly.
Compared with traditional single-view data, multi-view data are informative and can provide a more comprehensive description\cite{geng2021uncertainty,zhang2018generalized,zhang2019ae2,zheng2020feature}.
Thanks to these appealing properties, the research of multi-view learning attracts increasing attention, and one of the challenging branches is Unsupervised Multi-view Representation Learning (UMRL). 
The purpose of UMRL is to learn a unified representation containing consistent information and complementary information, which is usually obtained by mapping data from different views into a shared low-dimensional space\cite{xu2013survey,lu2019tensor,qu2020context}.
Therefore, the unified representation learned from multiple views could be easily exploited by on-shelf classic single-view algorithms for various downstream tasks effectively\cite{zhang2019ae2}.
Obviously, a naive way to achieve the goal of UMRL is feature concatenation, which concatenates different views directly to get a joint feature representation. However, since the specific statistical properties among different views are diverse, the feature concatenation strategy usually leads to negative performance\cite{xu2013survey, zhao2017multi, zheng2022graph}.

In recent years, many methods are proposed to address the problem of UMRL.
For example, CCA\cite{hotelling1992relations} and CCA-based methods\cite{akaho2006kernel, andrew2013deep, chen2019graph,wang2015deep} map different views into a low-dimensional space based on the canonical correlation analysis.
FMRL\cite{li2019flexible} learns the unified representation by utilizing the Hilbert-Schmidt independence criterion \cite{gretton2005measuring} to capture the non-linear correlations of multiple views.
AE$^2$-Nets\cite{zhang2019ae2} introduces the nested autoencoder networks to learn the unified feature representation.
DUA-Nets\cite{geng2021uncertainty} investigates the information of multiple views by employing uncertainty modeling and learns the noise-free feature representation.
Although gratifying progress is made and the promising unified multi-view representation can be learned by these aforementioned methods, they are all focused on fusing the multi-view information in the feature space while neglecting the important information in the semantic space.
Since different views describe the same object, it is more likely and reasonable to exploit the consensus in the semantic space rather than simply in the feature space during the fusion process\cite{xu2022multi, li2021contrastive}.
Compared to pursuing consensus in the feature space, exploring consensus in the semantic space can effectively preserve the diversity information of views during the learning process of UMRL.

\begin{figure*}[t] 
\centering
\includegraphics[width=1\textwidth]{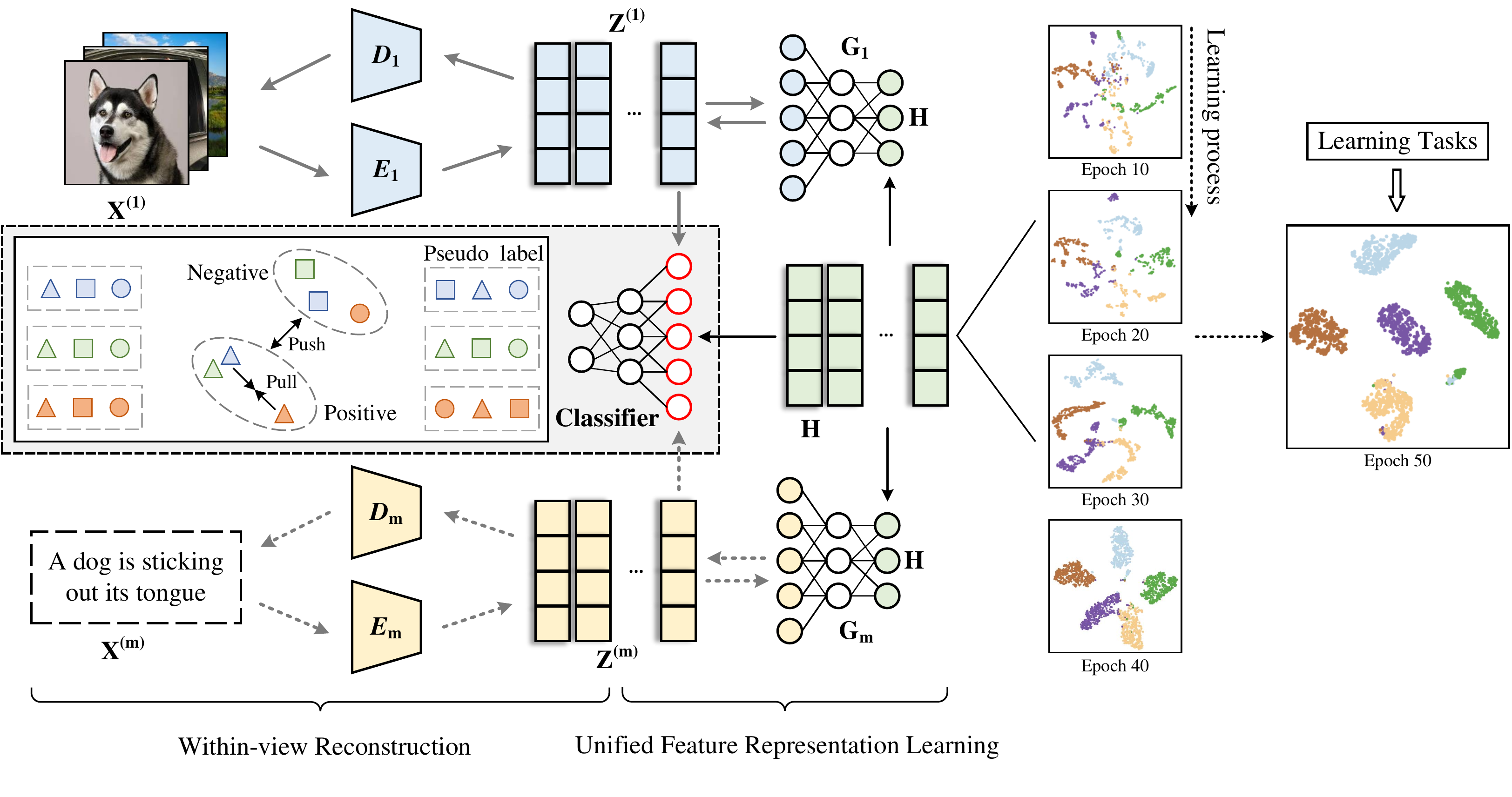}
\caption{
Overview of SCMRL, which explores and exploits the consensus in the semantic space to boost the learning performance of the unified feature representation of multi-view data.
More specifically, given multi-view data $\{\mathbf{X}^{(i)}\}_{i=1}^m$ with $m$ views, the within-view reconstruction modules of different views are used to obtain the view-specific feature representation $\{\mathbf{Z}^{(i)}\}_{i=1}^m$, and the unified feature representation learning module dynamically learns the unified feature representation $\mathbf{H}$. Meanwhile, the shared classification network and the contrastive learning are introduced to bridge these two modules so that semantic consistency information is captured by $\mathbf{H}$. On the right side, t-SNE is used to visualize the learned $\mathbf{H}$ during the learning process (visualization results are based on the BDGP dataset).} \label{fg1} 
\label{Networks} 
\end{figure*}

As we discussed above, it is promising to seek the multi-view consensus information in the semantic space to guide the learning process of UMRL.
To this end, we propose a novel method, named Semantically Consistent Multi-view Representation Learning (SCMRL), and the framework of SCMRL is depicted in Figure \ref{fg1}. %
Specifically, SCMRL has two basic modules, namely within-view reconstruction and unified feature representation learning, they are novelty integrated by the exploration of multi-view consensus information in the semantic space.

Different from most existing methods, such as AE$^2$-Nets \cite{zhang2019ae2} and CUMRL \cite{zheng2021collaborative}, which mainly consider multi-view information in the feature space, the proposed method employs the multi-view consensus information in the semantic space.
As shown in Figure~\ref{Networks}, we introduce the shared classification network to obtain the pseudo labels of both view-specific feature representations $\{\mathbf{Z}^{(i)}\}_{i=1}^m$ and the desired unified feature representation $\mathbf{H}$.
The shared classification network acts as an intermediary between within-view reconstruction and unified feature representation learning, which excavates valuable semantic information in the semantic space.
Since different views and the learned unified representation describe the same object, their pseudo-labels tend to be consistent.
To achieve this goal, we introduce contrastive learning to explore the multi-view consensus information in the semantic space and align the semantic labels of both view-specific feature representations and the learned unified feature representation simultaneously.
Based on the aforementioned learning process, the unified representation $\mathbf{H}$ can effectively integrate the information of multiple views in both feature space and semantic space.

The main contributions of the proposed method can be summarized as follows:

$\bullet$ We introduce a novel Semantically Consistent Multi-view Representation Learning (SCMRL), which learns the unified feature representation under the guidance of the consensus semantic information from multi-view data.

$\bullet$ The contrastive learning strategy is specially designed to bridge the within-view reconstruction and unified feature representation learning in SCMRL. The consensus information in the semantic space can be fully exploited to constrain the learning process of the unified feature representation.

$\bullet$ Extensive experiments are conducted, and experimental results verify the effectiveness of the proposed SCMRL compared with several state-of-the-art UMRL methods.

\section{Related Work}
In this section, we will review recent works of UMRL.
Since contrastive learning is also leveraged in our method, we will briefly introduce related multi-view learning works based on contrastive learning as well.

\subsection{\bf UMRL}
The goal of UMRL is to learn a promising representation of multi-view data without supervision.
The learned unified feature representation from multiple views can be straightforwardly leveraged by off-the-shelf classic and effective algorithms for downstream tasks, such as classification tasks, clustering tasks, and recognition tasks \cite{zhang2019ae2,wang2015deep,geng2021uncertainty,zheng2023comprehensive}.
Based on the advantage and the effectiveness of multi-view data, UMRL has attracted increasing attention, and many methods have been proposed in recent years \cite{li2021contrastive, liu2021multiview, huang2021deep}.

The representative methods are CCA-based\cite{akaho2006kernel,andrew2013deep,chen2019graph,wang2015deep} methods, which aim to maximize the canonical correlation among views in the low-dimensional space. Taking a multi-view dataset with two views for example here, CCA-based methods have the following basic formula:
\begin{equation}
\label{ep1}
    \min_{\beta _{1}, \beta _{2}} -\bm{I}(  \bm{f}_{1}( \mathbf{X}^{(1) }  ; \beta _{1}) , \bm{f}_{ 2}( \mathbf{X}^{(2) }  ; \beta _{2})) + \lambda \bm{g}( \beta _{1},\beta _{2}),
\end{equation}
where $\bm{f}_{1}(\cdot; \beta_{1})$ and $\bm{f}_{2}(\cdot; \beta_{2})$ are two embedding strategies with parameters $\beta _{1}$ and $\beta _{2}$. $\bm{I}(\cdot)$ and $\bm{g}(\cdot)$ indicate the canonical correlation function and the regularization term respectively. For example, DCCA \cite{andrew2013deep} utilizes deep neural networks for the reconstruction processes of different views. To make the learning process more reliable, DCCAE \cite{wang2015deep} considers the bottleneck representations by employing autoencoders to minimize the reconstruction loss.

In addition to these CCA-based methods, there are some other UMRL methods that have been proposed in recent years. For example, CMRL\cite{zheng2023comprehensive} fuses the low-dimensional embedding representations and imposes the low-rank tensor constraints on the subspace representations of different views to learn a unified feature representation with comprehensive information.
AE$^2$-Nets\cite{zhang2019ae2} is a learning framework consisting of nested autoencoders, which is designed to achieve the compact unified multi-view feature representation by balancing the complementarity and consistency properties among views.
MvLNet \cite{huang2021deep} can learn the unified multi-view spectral representation, and uses the Cholesky decomposition during the learning process by introducing the orthogonal constraint and reformulation strategy. 
DCP\cite{lin2022dual} is based on information theory, which maximizes mutual information of different views based on contrastive learning to achieve the goal of UMRL.

\subsection{\bf Contrastive Learning}
Contrastive learning\cite{chen2020simple,xie2021detco} is an effective representation learning method, which maximizes the similarity of positive pairs and minimizes the similarity of negative pairs in latent space.
This learning paradigm performs well in computer vision\cite{niu2021spice, van2020scan}.
It is also widely used in multi-view clustering in recent years\cite{li2021contrastive, lin2021completer, roy2021self, lin2022contrastive}.
SURE\cite{yang2022robust} has designed a contrastive learning loss specially used for incomplete multi-view clustering, which uses the available pairs as positive pairs and randomly selects some cross-view samples as negative pairs.
It effectively and robustly solves the partial view-unaligned problem (PVP) and partial sample-missing problem (PSP) in multi-view clustering.
MFLVC\cite{xu2022multi} optimizes different goals in different feature spaces through contrastive learning and solves the conflict between view reconstruction and consistency goals, which effectively learns common semantics and avoids the influence of meaningless view-specific information.

\section{Proposed Approach}
Given a multi-view dataset $\{\mathbf{X}^{(i)}\} _{i=1}^{m}$ with $m$ view, the $i$-th view in the original feature space is denoted by $\mathbf{X}^{(i)} \in \mathbf{R}^{N \times D_{i}}$, where $N$ denotes the number of samples and $D_{i}$ represents the dimension of the feature space. To learn the promising unified multi-view feature representation $\mathbf{H} \in \mathbf{R}^{N \times D_{H}}$, we propose a novel semantically consistent multi-view representation learning (SCMRL), 
the framework of which is shown in Figure \ref{fg1}.

\subsection{\bf SCMRL}
Overall, the loss function of SCMRL can be formulated as follows:
\begin{equation}
\label{ep2}
    \mathcal{L} = \mathcal{L}_{Rec} + \lambda _{1}\mathcal{L}_{Deg} +\lambda _{2}\mathcal{L}_{Sem},
\end{equation}
where $\mathcal{L}_{Rec}$ is the loss of within-view reconstruction, $\mathcal{L}_{Deg}$ indicates the loss of the degradation learning employed in unified feature representation learning, and $\mathcal{L}_{Sem}$ denotes the loss of semantic consistency. Regarding $\lambda _{1}$ and $\lambda _{2}$, they are two trade-off parameters.

\subsubsection{\bf Within-view Reconstruction}
Generally, the original multi-view data contains a lot of redundant and noisy information, which will have a negative impact on downstream tasks.
To simultaneously deal with multi-view data conveniently and learn a reliable representation of each view, we utilize deep autoencoders to obtain the view-specific feature representation of each view.
Specifically, $\bm{E}_{i}(\cdot;\theta _{i})$ and $\bm{D}_{i}(\cdot;\phi _{i})$ represent the encoder and decoder of the $i$-th view respectively, where $\theta _{i}$ and $\phi _{i}$ denote the corresponding parameters.

Subsequently, each view can be encoded into a low-dimensional feature as follows:
\begin{equation}
\label{ep3}
    \mathbf{Z}_{j}^{(i)}  =  \bm{E}_{i}(\mathbf{X}_{j}^{(i)};\theta _{i} ),
\end{equation}
where $\mathbf{X}_{j}^{(i)}$ is the $j$-th sample of $\mathbf{X}^{(i)}$, $\mathbf{Z}_{j}^{(i)} \in \mathbf{R}^{D_{Z}}$ denotes the embedded feature in the ${D_{Z}}$-dimensional feature space. Then we input this low-dimensional feature into the decoder for reconstruction:
\begin{equation}
\label{ep4}
    \mathbf{\hat{X}}_{j}^{(i)}  =  \bm{D}_{i}(\mathbf{Z}_{j}^{(i)};\phi  _{i} ),
\end{equation}
where $\mathbf{\hat{X}}_{j}^{(i)}$ is the reconstructed representation. Therefore, we can get the following reconstruction loss $\mathcal{L}_{Rec}$:
\begin{equation}
\label{ep5}
    \mathcal{L}_{Rec} = \sum_{i=1}^{m} \sum_{j=1}^{N} ||\mathbf{X}_{j}^{(i)}-\bm{D}_{i}(\bm{E}_{i}(\mathbf{X}_{j}^{(i)};\theta _{i});\phi  _{i}) ||_{2}^{2}. 
\end{equation}
By minimizing the reconstruction loss $\mathcal{L}_{Rec}$, we can transform the input $\mathbf{X}^{(i)}$ into the representation $\mathbf{Z}^{(i)}$.

\subsubsection{\bf Unified Feature Representation Learning}
Based on the within-view reconstruction, we can obtain the low-dimensional view-specific representations $\{\mathbf{Z}^{(i)}\}_{i=1}^m$ of different views. Since the desired unified feature representation should contain comprehensive information of multi-views, it is reasonable that different views require to be reconstructed into a unified feature representation. Considering the way of directly adding the low-dimensional view-specific representation of multiple views, it neglects the different importance and the diverse specific statistical properties of multiple views, and usually leads to poor performance. To effectively learn the unified feature representation $\mathbf{H}$, the degradation learning strategy is adopted in the proposed method. Specifically, we introduce the degradation learning process of the $i$-th view:
\begin{equation}
\label{ep6}
    \mathcal{L}_{Deg} = \sum_{i=1}^{m} \sum_{j=1}^{N} ||\mathbf{Z}_{j}^{(i)}-\bm{G}_{i}(\mathbf{H}_{j};\delta _{i})||_{2}^{2},
\end{equation}
where $\bm{G}_{i}(\cdot;\delta _{i})$ is the fully connected degradation neural network with parameter $\delta _{i}$ and $\mathbf{H}$ can be updated during the learning process. Based on Eq.~(\ref{ep6}), the unified feature representation learning can dynamically balance the importance of multiple views and integrate the low-dimensional view-specific representation.

To initialize the unified feature representation, the following strategy is employed: 
\begin{equation}
\label{ep7}
    \mathbf{H}_{j}  = \frac{1}{m}\sum_{i=1}^{m} (\mathbf{Z}_{j}^{(i)}),
\end{equation}
which can ensure that $\mathbf{H}$ contains comprehensive information of multi-views.

\subsubsection{\bf Contrastive learning of semantic consistency}
As we discussed above, we can observe that both the learning process of within-view reconstruction and unified feature representation learning exploit the multi-view information in the feature space. To effectively excavate the variable semantic consensus information in semantic space, the contrastive learning of semantic consistency is introduced here. According to the fact that multiple views and the unified representation describe the same objective, we introduce a shared classification network, termed $\bm{Classifer}(\cdot;\varphi)$ with the parameter $\varphi$. Naturally, we constrain that $\{\mathbf{Z}^{(i)}\}_{i=1}^m$ and $\mathbf{H}$ should have similar pseudo labels. By utilizing $\bm{Classifer}(\cdot;\varphi)$, we map the representations, including $\{\mathbf{Z}^{(i)}\}_{i=1}^m$ and $\mathbf{H}$, to the semantic space with dimension size of $k$, where $k$ is the number of categories of the multi-view dataset.

In other words, we have the following formula:
\begin{equation}
\label{ep8}
    \{\mathbf{Q}_{j}^{(i)}, \mathbf{Q}_{j}^{\mathbf{H}}\} = \bm{Classifer}(\{\mathbf{Z}_{j}^{(i)}, \mathbf{H}_j\};\varphi),
\end{equation}
in which $\mathbf{Q}_{j}^{(i)}$ and $\mathbf{Q}_{j}^{\mathbf{H}}$ denote the pseudo labels of the $\mathbf{Z}_{j}^{(i)}$ and the unified feature representation $\mathbf{H}_{j}$, respectively. For convenience, we denote $\mathbf{Q}_{j}^{(m+1)}= \mathbf{Q}_{j}^{\mathbf{H}}$. Besides,
$\mathbf{Q}_{j}^{(i)}$ is formulated as:
\begin{equation}
\label{ep9}
    \mathbf{Q}_{j}^{(i)} = [q_{j1}^{(i)},q_{j2}^{(i)}...q_{jk}^{(i)}],
\end{equation}
where $q_{jk}^{(i)}$ is the probability of the $j$-th sample in the $i$-th view belonging to the $k$-th class.

Due to the diverse specific statistic information of multiple views, different views may have confused semantic information in semantic space, which leads to various and confusing results of $\mathbf{Q}_{j}^{(i)}$ and $\mathbf{Q}_{j}^{\mathbf{H}}$.
Therefore, we introduce contrastive learning \cite{chen2020simple} to mine the consistent semantic information in the semantic space and obtain consistent categories simultaneously.

Specifically, the sematic column vector $q_{\cdot w}^{(i)}$ have $((m+1)k - 1)$ vector pairs $\{q_{\cdot w}^{(i)}, q_{\cdot c}^{(j)}\}_{c=1,...,k}^{j=1,...,m+1}$, which contain $m$ positive pairs $\{q_{\cdot w}^{(i)}, q_{\cdot w}^{(j)}\}_{j \neq i}$ and the rest $(k-1)(m+1)$ negative pairs.
The cosine similarity is utilized to measure the similarity between two semantic column vectors:
\begin{equation}
\label{ep10}
    cos(q_{\cdot c}^{(i)}, q_{\cdot w}^{(j)}) = \frac{q_{\cdot c}^{(i)} \cdot q_{\cdot w}^{(j)}}{||q_{\cdot c}^{(i)}||||q_{\cdot w}^{(j)}||}.  
\end{equation}

\begin{algorithm}[t]
	\caption{Optimization algorithm of SCMRL}
	\label{alg1}
	\KwIn{Multi-view dataset $\{\mathbf{X}^{(i)}\} _{i=1}^{m}$; Parameter $\tau$; The number of categories k.}
	\KwOut{Unified Feature Representation $\mathbf{H}$.}  
	\BlankLine
	Initialize $\{\theta _{i}, \phi _{i}\}_{i=1}^m$ by minimizing Eq.~(\ref{ep5});
	
	Initialize $\mathbf{H}$ by Eq.~(\ref{ep7});
	
	\While{\textnormal{not converged}}{
		Obtain the view-specific representation $\{\mathbf{Z}^{(i)}\}_{i=1}^m$ through Eq.~(\ref{ep3});
		
		Obtain the label distribution of each view and the unified feature representation by Eq.~(\ref{ep8});
		
		Update $\{\theta _{i}, \phi _{i} ,\delta _{i}\}_{i=1}^m$, $\mathbf{H}$ and $\varphi$ with Eq.~(\ref{ep2});
	}
\end{algorithm}

Then, contrastive loss $l_{two}(i, j)$ between semantic column vectors $q_{\cdot c}^{(i)}$ and $q_{\cdot c}^{(j)}$ is defined as:
\begin{equation}
\label{ep11}
     l_{two}(i,j)=- \sum_{c=1}^{k}log(f(i,j,c)), 
\end{equation}
where
\begin{equation}
   f(i,j,c) = \frac{e^{cos(q_{\cdot c}^{(i)},q_{\cdot c}^{(j)})/\tau } }{ (\sum_{w=1}^{k}(e^{cos(q_{\cdot c}^{(i)}, q_{\cdot w}^{(i)})/\tau }+e^{cos(q_{\cdot c}^{(i)}, q_{\cdot w}^{(j)})/\tau })-e^{\frac{1}{\tau} }} \nonumber
\end{equation}
and $\tau$ denotes the temperature parameter. Further, it is convenient to define the complete comparative learning loss of semantic consistency $L_{Sem}$ as:
\begin{equation}
\label{ep12}
    L_{Sem} = l_{Sum} + l_{Reg}, 
\end{equation}
where $l_{Sum}$ denotes the contrastive loss for the whole dataset and $l_{Reg}$ indicates the regularization phase. Specifically, the item $l_{Sum}$ is defined as:
\begin{equation}
\label{ep13}
    l_{Sum} = \frac{1}{2}  \sum_{i=1}^{m+1} \sum_{j=1, j \neq i}^{m+1} \frac{l_{two}(i, j)}{k}. 
\end{equation}
And the item $l_{Reg}$ is formulated as:
\begin{equation}
\label{ep14}
    l_{Reg} = \sum_{i=1}^{m+1}\sum_{c=1}^{k} (\frac{1}{N}\sum_{j=1}^{N}q_{jc}^{(i)} \log_{}{\frac{1}{N}\sum_{j=1}^{N}q_{jc}^{(i)}}  ), 
\end{equation}
which can avoid grouping all samples into the same cluster.

For clarification, the optimization procedure of SCMRL is summarized in Algorithm \ref{alg1}.

\section{Experiments}
To verify the effectiveness of our method, extensive experiments are conducted in this section. Specifically, two basic tasks, namely clustering and classification, are used to evaluate the performance of the learned unified multi-view feature representation. Furthermore, detailed discussions of our method are provided as well.

\subsection{\bf Experiments Setup}

\begin{table*}[]
\caption{Details of the used datasets.} 
\label{table_data}
\resizebox{1\textwidth}{!}{
\begin{tabular}{ccccc}
\toprule
Dataset    & \#Sample & \#Cluster & \#View & \#Dimensionality of features  \\ \midrule
MNIST-USPS & 5000     & 10        & 2      & \{1*28*28, 1*28*28\}          \\
BDGP       & 2500     & 5         & 2      & \{1750, 79\}                  \\
Fashion    & 10000    & 10        & 3      & \{1*28*28, 1*28*28, 1*28*28\} \\
CCV        & 6773     & 20        & 3      & \{5000, 5000, 4000\}          \\
Caltech-2V & 1400     & 7         & 2      & \{40, 254\}                   \\
Caltech-3V & 1400     & 7         & 3      & \{40, 254, 928\}              \\
Caltech-4V & 1400     & 7         & 4      & \{40, 254, 928, 512\}         \\
Caltech-5V & 1400     & 7         & 5      & \{40, 254, 928, 512, 1984\}   \\ \bottomrule
\end{tabular}}
\end{table*}

\subsubsection{\bf Datasets}
We use the following benchmark datasets, the detail of these datasets are shown in Table \ref{table_data}:

{\bf 1) MNIST-USPS}\cite{peng2019comic}: It is a two-view dataset that contains 5000 handwritten digital image samples from numbers 0 to 9.

{\bf 2) BDGP}\cite{cai2012joint}: It is a two-view dataset containing 2500 images of drosophila embryos belonging to 5 categories.

{\bf 3) Fashion}\cite{xiao2017fashion}: It has 10000 images collected from 10 categories about fashion products and has three views.

{\bf 4) CCV}\cite{jiang2011consumer}: It contains 6773 internet videos samples belonging to 20 classes.
It has three views, such as STIP, SIFT, and MFCC.

{\bf 5) Caltech}\cite{fei2004learning}: It is collected from 1400 images, which belong to 7 categories and have five views.
Four sub-datasets, namely Caltech-2V, Caltech-3V, Caltech-4V, and Caltech-5V, with different numbers of views are built for evaluation here. Specifically, Caltech-2V uses WM and CENTRISTT; Caltech-3V uses WM, CENTRIST, and LBP; Caltech-4V uses WM, CENTRIST, LBP, and GIST; Caltech-5V uses WM, CENTRIST, LBP, GIST, and HOG.

\subsubsection{\bf Comparison methods}
The following state-of-the-arts are used for comparison:

{\bf 1) DCCA}\cite{andrew2013deep}: It is a classic CCA-based method, which uses depth neural networks to obtain the nonlinear mapping with the maximum view linear correlation.

{\bf 2) DCCAE}\cite{wang2015deep}: It is also a CCA-based method. Different from DCCA, it uses autoencoders to obtain the low dimensional projection of the original data and maximizes the view correlation between the learned representations.

{\bf 3) LMSC\cite{zhang2017latent}}: It learns the latent data representation by mapping different views into the common space and employing the low-rank subspace constraint.

{\bf 4) AE$^2$-Nets}\cite{zhang2019ae2}: Nested autoencoders are used to learn the compact unified representation by balancing the complementarity and consistency among multi-views.

{\bf 5) DUA-Nets}\cite{geng2021uncertainty}: It presents the dynamic uncertainty-aware networks for UMRL.
By estimating and leveraging the uncertainty of data, it achieves the noise-free multi-view feature representation.

{\bf 6) CUMRL}\cite{zheng2021collaborative}: It considers the low-rank tensor constraint to excavate the high-order view correlations of multi-view data in the feature space, and introduces the collaborative learning strategy for UMRL.

{\bf 7) DCP}\cite{lin2022dual}: It learns the unified multi-view representation by maximizing the mutual information of different views via contrastive learning in the feature space.

{\bf 8) CMRL}\cite{zheng2023comprehensive}: It achieves the unified multi-view representation with comprehensive information by introducing the orthogonal mapping strategy and imposing the low-rank tensor constraint on the subspace representations.

\subsubsection{\bf Evaluation metrics} 
Two basic tasks, i.e., clustering and classification, are employed in this section. Since $k$-means and $k$-nearest neighbor ($k$NN) are simple and can effectively measure the quality of the learned unified representation, we adopt the classic $k$-means algorithm for the clustering task and $k$NN algorithm for the classification task here\cite{zhang2019ae2,zheng2022graph,geng2021uncertainty}.

{\bf 1) Clustering task}: Three metrics are utilized to evaluate the clustering quality, namely ACC, Normalized Mutual Information (NMI), and Fscore.
For each dataset, 10 trials are conducted for all experiments to eliminate the randomness and make the experimental results more reliable.

{\bf 2) Classification task}: We utilize ACC as the metric to evaluate the classification performance. Specifically, we set k=5 for $k$NN in all experiments and 30 trials are conducted for each experiment.
We divide samples of the learned representation into training and testing sets with different proportions, and the ratios (\#Train/\#Test) are set to 8/2 (80\%/20\%), 5/5 (50\%/50\%), 2/8 (20\%/80\%).

\begin{figure*}[t]
	\centering
	\subfigure[Epoch 10]{
        \includegraphics[width=0.29\linewidth]{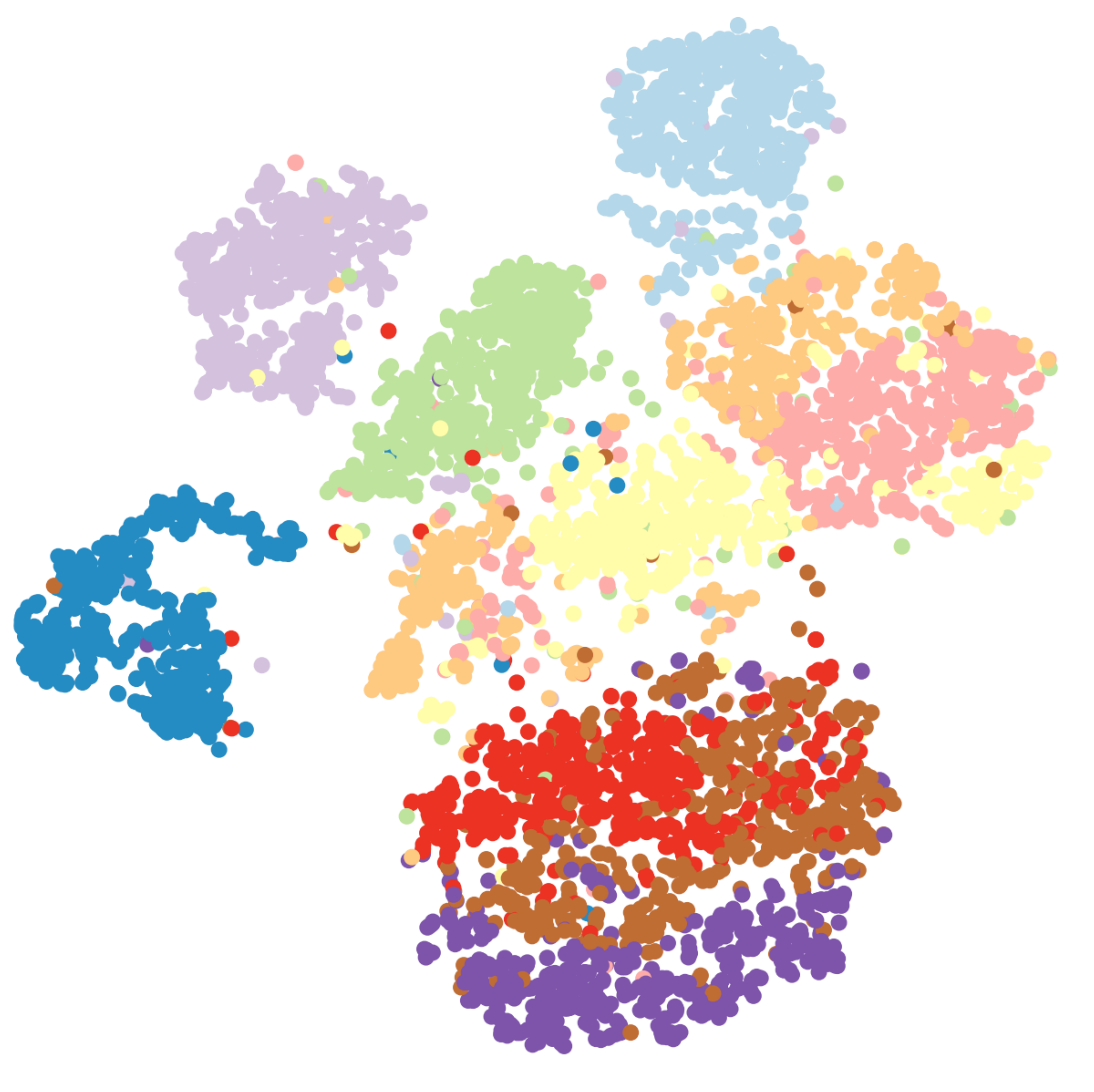}
        }
        \quad
        \subfigure[Epoch 20]{
        \includegraphics[width=0.29\linewidth]{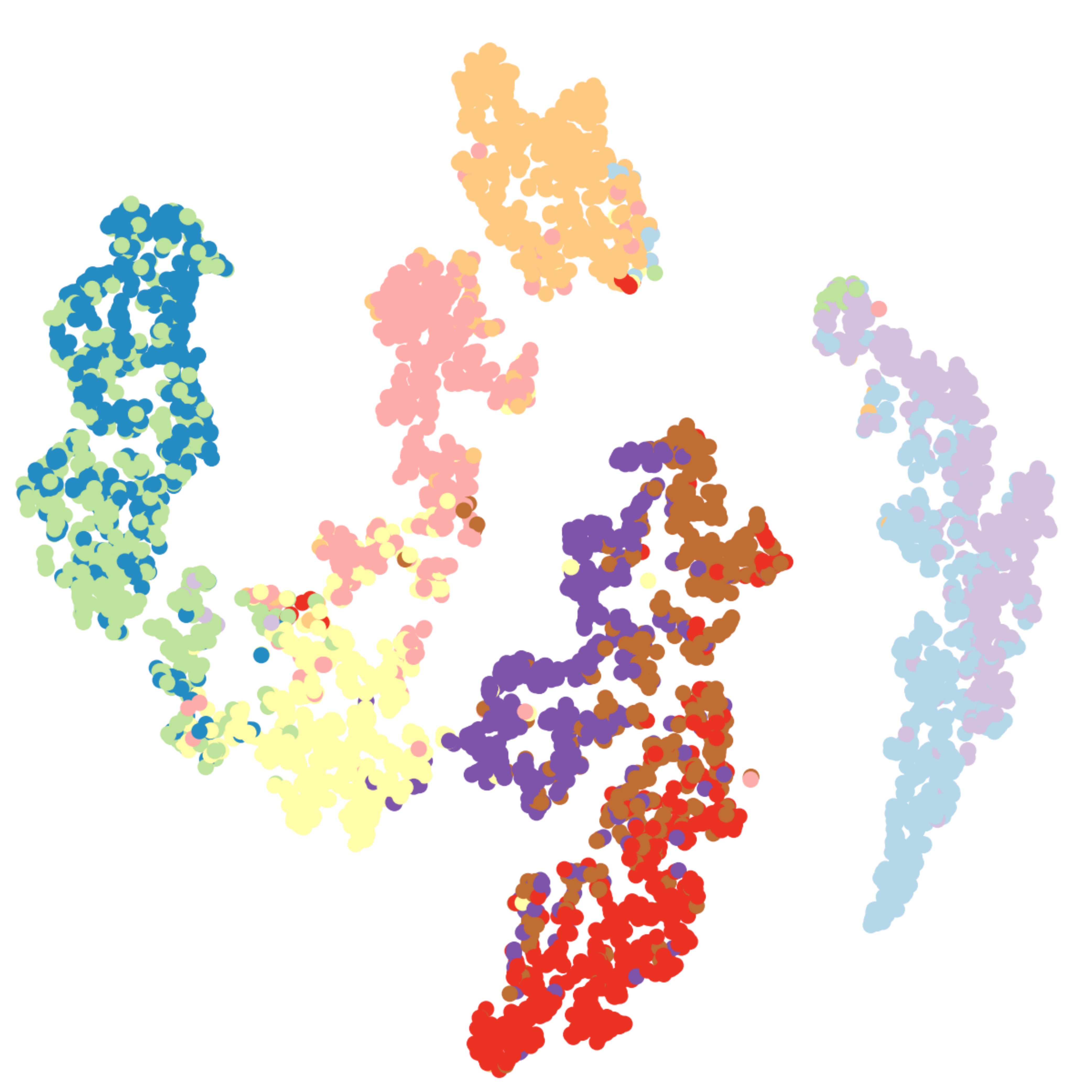}
        }
        \quad
        \subfigure[Epoch 30]{
        \includegraphics[width=0.29\linewidth]{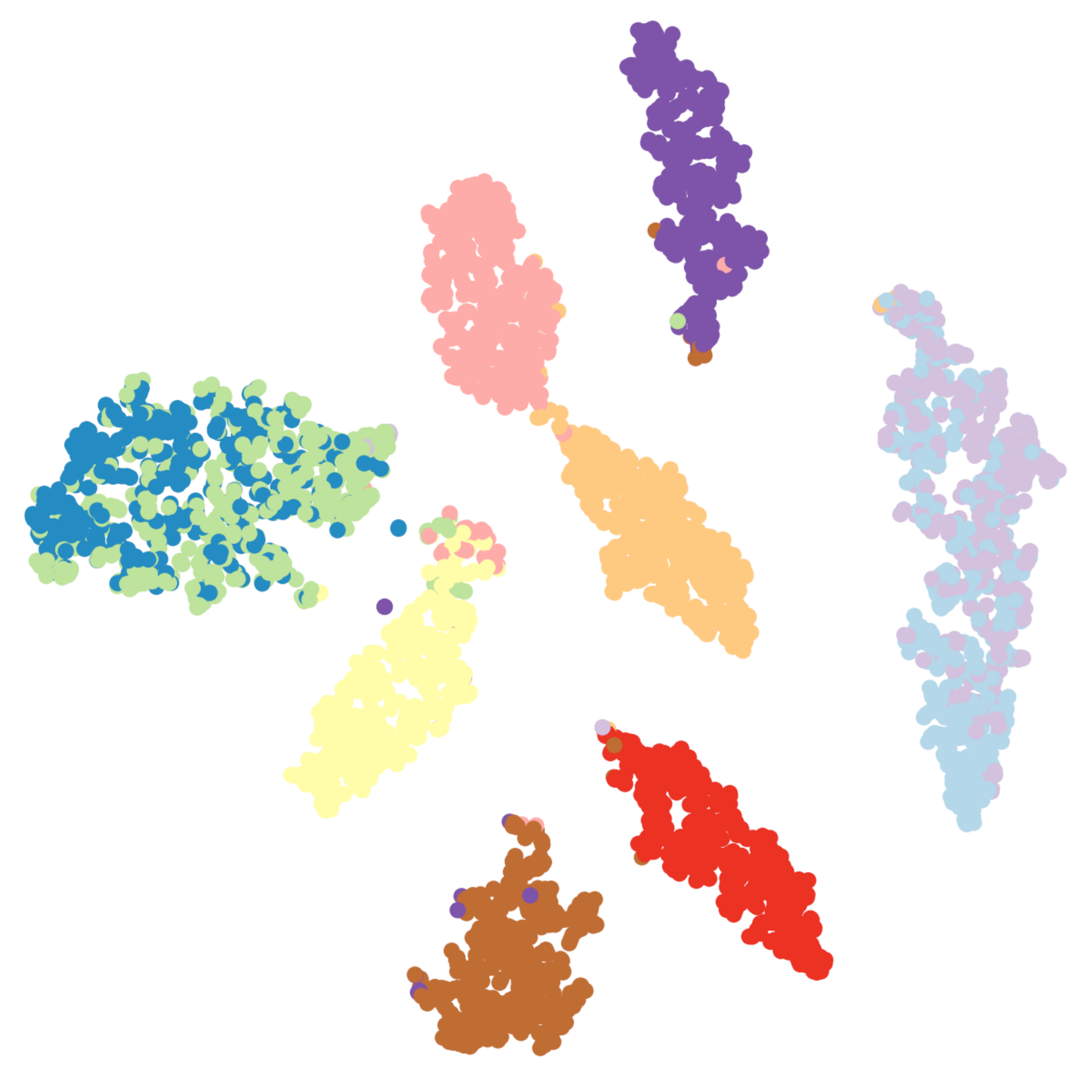}
        }
        \quad
        \subfigure[Epoch 40]{
        \includegraphics[width=0.28\linewidth]{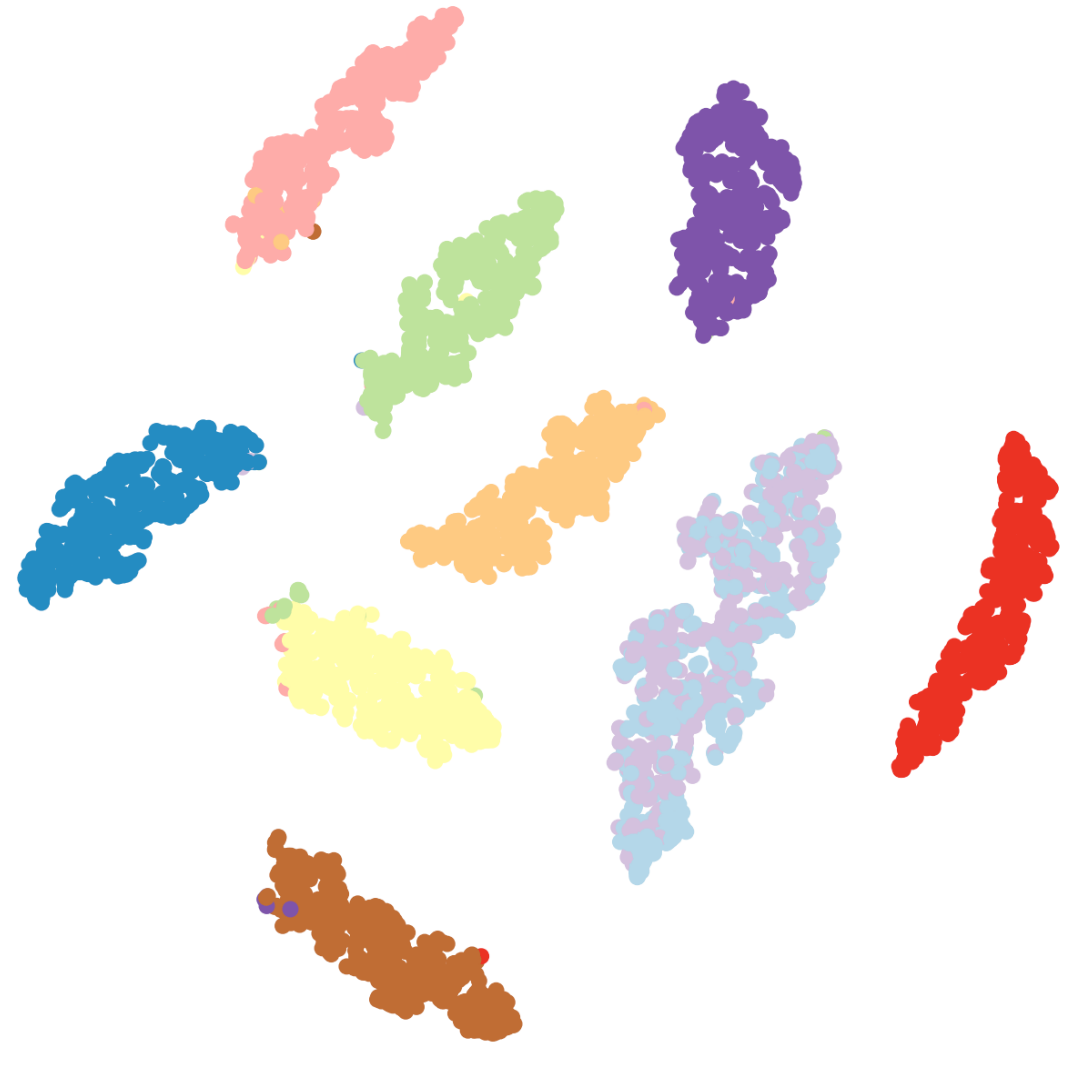}
        }
        \quad
        \subfigure[Epoch 50]{
        \includegraphics[width=0.28\linewidth]{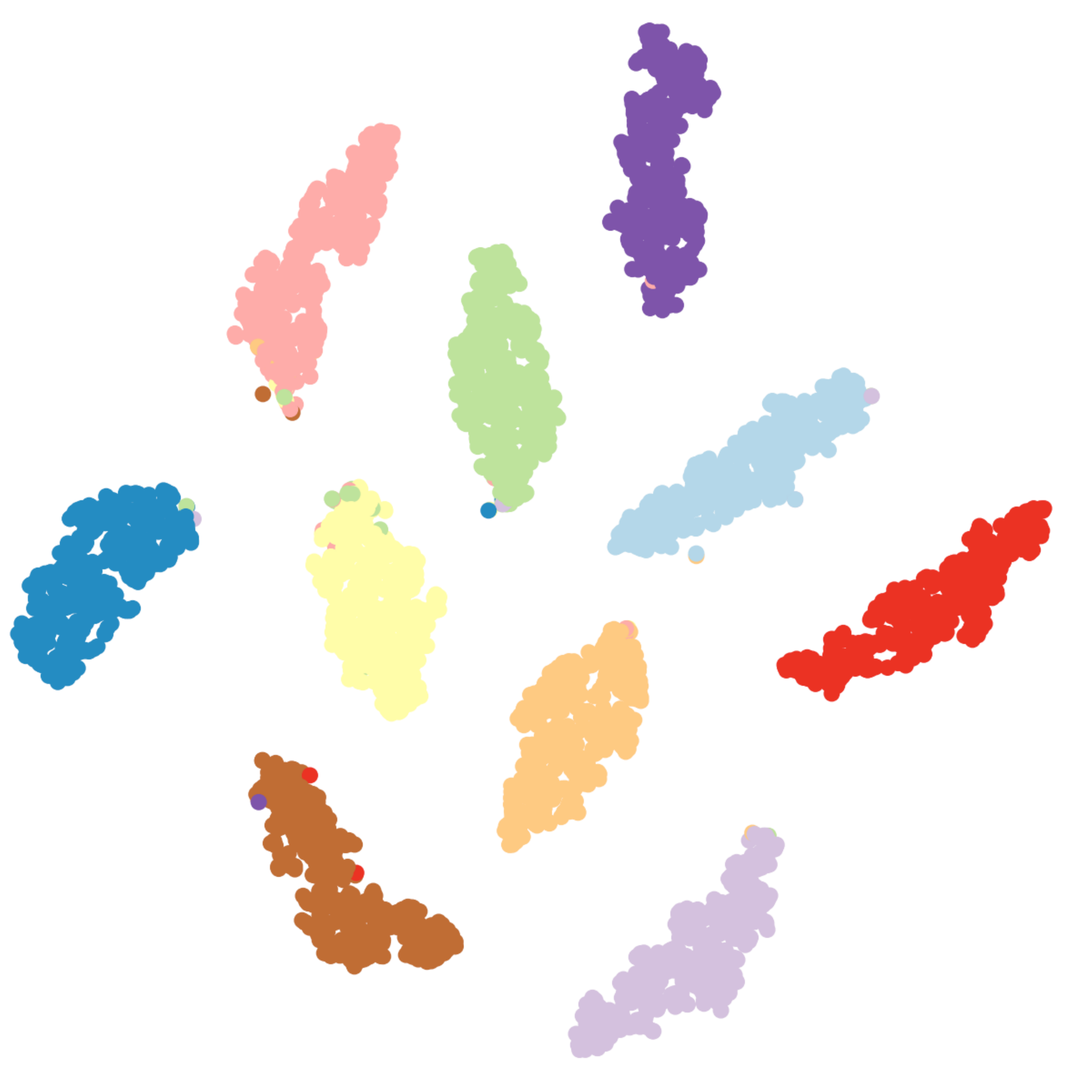}
        }
        \quad
	\caption{The t-SNE visualization results of the learned unified feature representation in Epoch 10, 20, 30, 40, and 50 of the learning process. Experimental results on the MNIST-USPS dataset are shown here.}
	\label{fgg2}
\end{figure*}

\begin{table*}[t]
\centering  
\caption{Clustering results of the learned unified multi-view representation on BDGP, MNIST-USPS, Fashion and CCV.}  
\label{table1.}
\resizebox{1\textwidth}{!}{
\begin{tabular}{rll|ccc|ccc|ccc|ccc}
\toprule
\multicolumn{3}{r|}{Datasets} & \multicolumn{3}{c|}{BDGP}                & \multicolumn{3}{c|}{MNIST-USPS}          & \multicolumn{3}{c|}{Fashion}             & \multicolumn{3}{c}{CCV}                  \\ 
\multicolumn{3}{r|}{Metrics}   & ACC            & NMI            & Fscore & ACC            & NMI            & Fscore & ACC            & NMI            & Fscore & ACC            & NMI            & Fscore \\ \midrule
\multicolumn{3}{r|}{DCCA}     & 67.72          & 56.45          & 63.05  & 41.55          & 35.13          & 31.52  & 67.38          & 77.51          & 72.66  & 21.09          & \underline{19.14}          & 13.26  \\
\multicolumn{3}{r|}{DCCAE}    & 69.03               & 58.12               & 63.31       & 67.94               & 53.62               & 51.60       & 67.74               & 77.26               & 72.32       & 20.42               & 18.56               & 13.00       \\
\multicolumn{3}{r|}{LMSC}     & 52.39          & 43.15          & 55.63  & 37.25          & 43.33          & 40.42  & 43.90          & 41.68          & 36.71  & 13.49          & 8.660          & 11.76  \\
\multicolumn{3}{r|}{AE$^2$-Nets} & 55.24          & 40.57          & 50.13  & 62.59          & 62.31          & 56.71  & 72.94          & 76.27          & 71.57  & 9.430          & 2.810          & 7.590  \\
\multicolumn{3}{r|}{DUA-Nets} & 60.28          & 40.61          & 53.91  & 75.10          & 68.94          & 66.01  & 77.21          & 76.08          & 72.72  & 16.11          & 11.80          & 11.64  \\
\multicolumn{3}{r|}{CUMRL}    & 62.93          & 48.80          & 57.44  & 58.64          & 56.93          & 50.96  & 67.00          & 66.77          & 61.26  & 10.41          & 5.190          & 7.480  \\
\multicolumn{3}{r|}{DCP}      & 43.77          & 38.50          & 53.49  & 89.10          & \underline{94.13}          & \underline{92.87}  & 75.68          & \underline{86.19}          & \underline{82.23}  & 14.23          & 11.48          & 10.69  \\
\multicolumn{3}{r|}{CMRL}      & \underline{78.92}          & \underline{67.20}          & \underline{71.68}  & \underline{91.61}          & 85.60          & 85.98  & \underline{76.81}          & 80.29          & 74.77  & \underline{24.31}          & 18.11          & \underline{13.64}  \\
\multicolumn{3}{r|}{Ours}     & \textbf{98.00} & \textbf{94.69} & \textbf{96.18}       & \textbf{99.56} & \textbf{98.72} & \textbf{99.12}       & \textbf{98.90} & \textbf{97.32} & \textbf{97.84}       & \textbf{26.82} & \textbf{27.55} & \textbf{21.39}       \\ \bottomrule
\end{tabular}}
\end{table*}
\begin{table*}[t]
\centering  
\caption{Clustering results of the learned unified multi-view representation on Caltech-$n$V, $n$ is selected from $\{2,3,4,5\}$.}  
\label{table2.}
\resizebox{1\textwidth}{!}{
\begin{tabular}{rll|ccc|ccc|ccc|ccc}
\toprule
\multicolumn{3}{r|}{Datasets} & \multicolumn{3}{c|}{Caltech-2V}                  & \multicolumn{3}{c|}{Caltech-3V}                  & \multicolumn{3}{c|}{Caltech-4V}                  & \multicolumn{3}{c}{Caltech-5V}                   \\
\multicolumn{3}{r|}{Metrics}   & ACC            & NMI            & Fscore         & ACC            & NMI            & Fscore         & ACC            & NMI            & Fscore         & ACC            & NMI            & Fscore         \\ \midrule
\multicolumn{3}{r|}{DCCA}     & 39.36          & 30.69          & 37.02          & 46.71          & 35.29          & 37.78          & 54.97          & 33.44          & 38.53          & 62.73          & 42.97          & 48.13          \\
\multicolumn{3}{r|}{DCCAE}    & 43.87               & 31.17               & 36.23               & \underline{59.49}               & 44.53               & 47.15               & 50.29               & 32.01               & 37.78               & 63.70               & 45.58               & 49.62               \\
\multicolumn{3}{r|}{LMSC}     & 42.64          & 32.35          & 35.47          & 26.44          & 7.580          & 20.32          & 38.08          & 28.82          & 34.43          & 66.15          & 53.38          & 56.83          \\
\multicolumn{3}{r|}{AE$^2$-Nets} & 46.14          & 32.01          & 35.60          & 51.48          & 41.08          & 42.89          & 48.01          & 38.89          & 41.93          & 67.67          & 58.13          & 57.96          \\
\multicolumn{3}{r|}{DUA-Nets} & 39.45          & 22.12          & 30.88          & 43.63          & 29.37          & 36.96          & 46.31          & 34.54          & 41.03          & 56.89          & 44.37          & 48.46          \\
\multicolumn{3}{r|}{CUMRL}    & 48.74         & \underline{41.32}          & 42.36          & 56.59          & \underline{48.86}          & 49.29          & 67.76          & \underline{57.66}          & \underline{57.81}          & \underline{88.55}          & \underline{79.02}          & \underline{81.12}          \\
\multicolumn{3}{r|}{DCP}      & 42.64          & 32.35          & 35.47          & 51.60          & 48.35          & \underline{52.16}          & 53.37          & 53.55          & 55.54          & 54.04          & 54.05          & 56.16          \\
\multicolumn{3}{r|}{CMRL}      & \underline{55.03}          & 40.33          & \underline{42.95}          & 59.27          & 44.67          & 46.72          & \underline{69.30}          & 56.24          & 57.75          & 68.76          & 56.37          & 57.61          \\
\multicolumn{3}{r|}{Ours}     & \textbf{61.29} & \textbf{48.29} & \textbf{48.75} & \textbf{78.21} & \textbf{70.18} & \textbf{70.30} & \textbf{87.00} & \textbf{79.83} & \textbf{79.22} & \textbf{89.00} & \textbf{80.23} & \textbf{81.22} \\ \bottomrule
\end{tabular}}
\end{table*}

\subsubsection{\bf Implementation details}
For all datasets, the ReLU\cite{glorot2011deep} activation function is used to implement autoencoders in SCMRL.
Adam optimizer\cite{kingma2014adam} is employed for optimization.
Our method is implemented by PyTorch\cite{paszke2019pytorch} on one NVIDIA Geforce GTX 2080ti GPU with 11GB memory. For comparison methods, we leverage the codes released by their corresponding authors and use the recommended settings in their original works.

\subsection{\bf Visualization results}
To vividly reveal the structure of the learned unified representation, we visualize $\mathbf{H}$ achieved in the Epoch 10, 20, 30, 40, and 50 of the SCMRL learning process based on the t-SNE \cite{van2008visualizing}.
Taking the MNIST-USPS dataset for example here, the visualization results are shown in Figure \ref{fgg2}. It can be observed that the unified feature representation with a promising structure can be achieved by our method.

\subsection{\bf Experimental Results}
We discuss the experimental results of the clustering task and the classification task. Overall speaking, the proposed method can achieve the best performance in most cases.

\subsubsection{\bf Experimental results of the clustering task}
The $k$-means clustering results on all datasets are shown in Table \ref{table1.} and \ref{table2.}.
We can observe that SCMRL achieves the best performance on all datasets in all metrics and considerable progress can be made for all metrics. For example, on the BDGP dataset, our method respectively achieves an improvement of around 19.08\%, 27.49\%, and 24.50\% compared with the second-best results in metrics of ACC, NMI, and Fscore. Compared with the second-best results, 17.70\%, 22.17\%, and 21.41\% improvements can be obtained on the Caltech-4V in metrics of ACC, NMI, and Fscore, respectively. The main reason is that the proposed SCMRL explores the underlying consistent information of both the learned unified representation and multiple views in semantic space. Compared with other methods, our method can explore the consistent information in semantic space.
Consequently, the multi-view diversity information in the feature space can be preserved and utilized during the learning process of the unified feature representation. In Table \ref{table2.}, we can find that the performance of SCMRL steadily increases with the increase of views on Caltech-$n$V, which indicates that the rich information contained in multiple views can be effectively excavated and integrated into the learned unified representation by our method.

\begin{table*}[t]
\centering  
\caption{Classification results of the learned unified multi-view representation on BDGP, MNIST-USPS, Fashion and CCV.}  
\label{table3.}
\resizebox{1\textwidth}{!}{
\begin{tabular}{rll|ccc|ccc|ccc|ccc}
\toprule
\multicolumn{3}{r|}{Datasets} & \multicolumn{3}{c|}{BDGP}                        & \multicolumn{3}{c|}{MNIST-USPS}                  & \multicolumn{3}{c|}{Fashion}                     & \multicolumn{3}{c}{CCV}                          \\
\multicolumn{3}{r|}{\#Train/\#Test}   & 8/2            & 5/5            & 2/8            & 8/2            & 5/5            & 2/8            & 8/2            & 5/5            & 2/8            & 8/2            & 5/5            & 2/8            \\ \midrule
\multicolumn{3}{r|}{DCCA}     & 96.52          & 96.44          & 95.25          & 78.48          & 76.46          & 72.07          & 86.93          & 86.47          & 85.56          & 28.49          & 26.64          & 24.01          \\
\multicolumn{3}{r|}{DCCAE}    & 97.73               & 97.40               & 96.62               & 83.45               & 81.74               & 79.44               & 86.63               & 86.11               & 85.24               & 29.71               & 27.97               & 24.87               \\
\multicolumn{3}{r|}{LMSC}     & 98.16          & 97.60          & 96.22          & 94.66          & 93.34          & 89.55          & 86.12          & 84.80          & 81.27          & 33.77          & 31.17          & 26.02          \\
\multicolumn{3}{r|}{AE$^2$-Nets} & 91.65          & 91.19          & 87.56          & 96.43          & 95.62          & 93.30          & 91.83          & 90.95          & 89.10          & 6.380          & 6.280          & 6.070          \\
\multicolumn{3}{r|}{DUA-Nets} & 93.01          & 91.52          & 85.42          & 91.79          & 90.18          & 85.64          & 87.60          & 86.53          & 81.47          & 30.43          & 31.04          & 24.86          \\
\multicolumn{3}{r|}{CUMRL}    & 94.73          & 93.57          & 90.91          & 94.69          & 93.99          & 91.16          & 81.49          & 80.61          & 79.26          & 6.75           & 6.67           & 6.49           \\
\multicolumn{3}{r|}{DCP}      & 96.38          & 96.04          & 93.94          & \underline{98.49}          & \underline{98.33}          & \underline{97.95}          & 93.60          & 93.40          & \underline{92.44}          & 20.25          & 18.45          & 15.64          \\
\multicolumn{3}{r|}{CMRL}    & \underline{98.82}          & \underline{98.57}          & \underline{98.24}          & 97.86          & 97.45          & 96.28          & \underline{93.85}          & \underline{93.48}          & 92.36          & \underline{36.03}           & \underline{34.18}           & \underline{30.63}           \\
\multicolumn{3}{r|}{Ours}     & \textbf{98.92} & \textbf{98.74} & \textbf{98.54} & \textbf{99.63} & \textbf{99.57} & \textbf{99.57} & \textbf{99.11} & \textbf{99.08} & \textbf{99.04} & \textbf{39.41} & \textbf{37.85} & \textbf{35.09} \\ \bottomrule
\end{tabular}}
\end{table*}

\begin{table*}[t]
\centering  
\caption{Classification results of the learned unified multi-view representation on Caltech-$n$V, $n$ is selected from $\{2,3,4,5\}$.}  
\label{table4.}
\resizebox{1\textwidth}{!}{
\begin{tabular}{rll|ccc|ccc|ccc|ccc}
\toprule
\multicolumn{3}{r|}{Datasets} & \multicolumn{3}{c|}{Caltech-2V}                  & \multicolumn{3}{c|}{Caltech-3V}                  & \multicolumn{3}{c|}{Caltech-4V}                  & \multicolumn{3}{c}{Caltech-5V}                   \\
\multicolumn{3}{r|}{\#Train/\#Test}   & 8/2            & 5/5            & 2/8            & 8/2            & 5/5            & 2/8            & 8/2            & 5/5            & 2/8            & 8/2            & 5/5            & 2/8            \\ \midrule
\multicolumn{3}{r|}{DCCA}     & 68.24          & 65.41          & 60.61          & 72.81          & 71.14          & 66.81          & 65.80          & 64.26          & 61.04          & 62.73          & 42.97          & 48.13          \\
\multicolumn{3}{r|}{DCCAE}    & 69.82               & 67.12               & 62.92               & 74.46               & 73.01               & 68.23               & 66.80               & 64.75               & 60.19               & 63.70               & 45.58               & 49.62               \\
\multicolumn{3}{r|}{LMSC}     & 72.11          & 70.22          & 63.29          & 72.25          & 69.44          & 60.55          & 76.71          & 73.81          & 67.01          & 90.89          & 89.87          & 86.31          \\
\multicolumn{3}{r|}{AE$^2$-Nets} & 69.80          & 66.78          & 61.09          & 82.93          & 80.89          & 76.27          & 84.83          & 84.19          & 79.34          & 91.32          & 90.58          & 88.04          \\
\multicolumn{3}{r|}{DUA-Nets} & 62.73          & 59.40          & 54.37          & 69.58          & 69.28          & 63.05          & 75.88          & 74.29          & 66.69          & 81.04          & 80.29          & 77.51          \\
\multicolumn{3}{r|}{CUMRL}    & \underline{80.21} & \underline{77.64} & \underline{71.89} & \textbf{87.18} & \textbf{85.83} & \underline{81.62}          & \underline{90.31}          & \underline{89.70}          & \underline{86.59}          & \underline{93.48}          & \underline{93.29}          & \underline{91.83}          \\
\multicolumn{3}{r|}{DCP}      & 71.36          & 70.68          & 67.23          & 80.75          & 79.41          & 77.15          & 82.74          & 81.33          & 78.04          & 87.81          & 86.21          & 83.46          \\
\multicolumn{3}{r|}{CMRL}      & \textbf{80.81}          & \textbf{79.05}          & \textbf{74.10}          & 84.74          & 82.99          & 79.46          & 90.27          & 89.21          & 86.23          & 91.01          & 89.90          & 86.71          \\
\multicolumn{3}{r|}{Ours}     & 73.80          & 71.90          & 68.27          & \underline{84.82}          & \underline{83.30}          & \textbf{81.65} & \textbf{91.07} & \textbf{90.46} & \textbf{88.78} & \textbf{94.15} & \textbf{93.95} & \textbf{91.93} \\ \bottomrule
\end{tabular}}
\end{table*}

\subsubsection{\bf Experimental results of the classification task}
The classification results are shown in Table \ref{table3.} and Table \ref{table4.}. In general, the proposed method can obtain promising results for all datasets and achieve the best results for most cases. For example, on the Fashion dataset, our method can achieve an improvement of around 7\% with respect to the metric of ACC, compared with the second-best results.
Although CMRL and CUMRL achieve slightly better classification results on Caltech-2V and Caltech-3V, the results of our method are also competitive. Furthermore, with the increase of view on Caltech-$n$V, the performance of SCMRL can improve more significantly than CMRL and CUMRL. With the decrease of \#Train/\#Test, it can be observed that our method has the slowest performance decline, which also indicates the effectiveness of the semantic consistent information exploration in our method.

\begin{figure*}[t]
	\centering
	\subfigure[ACC in classification task]{
        \includegraphics[width=0.45\linewidth]{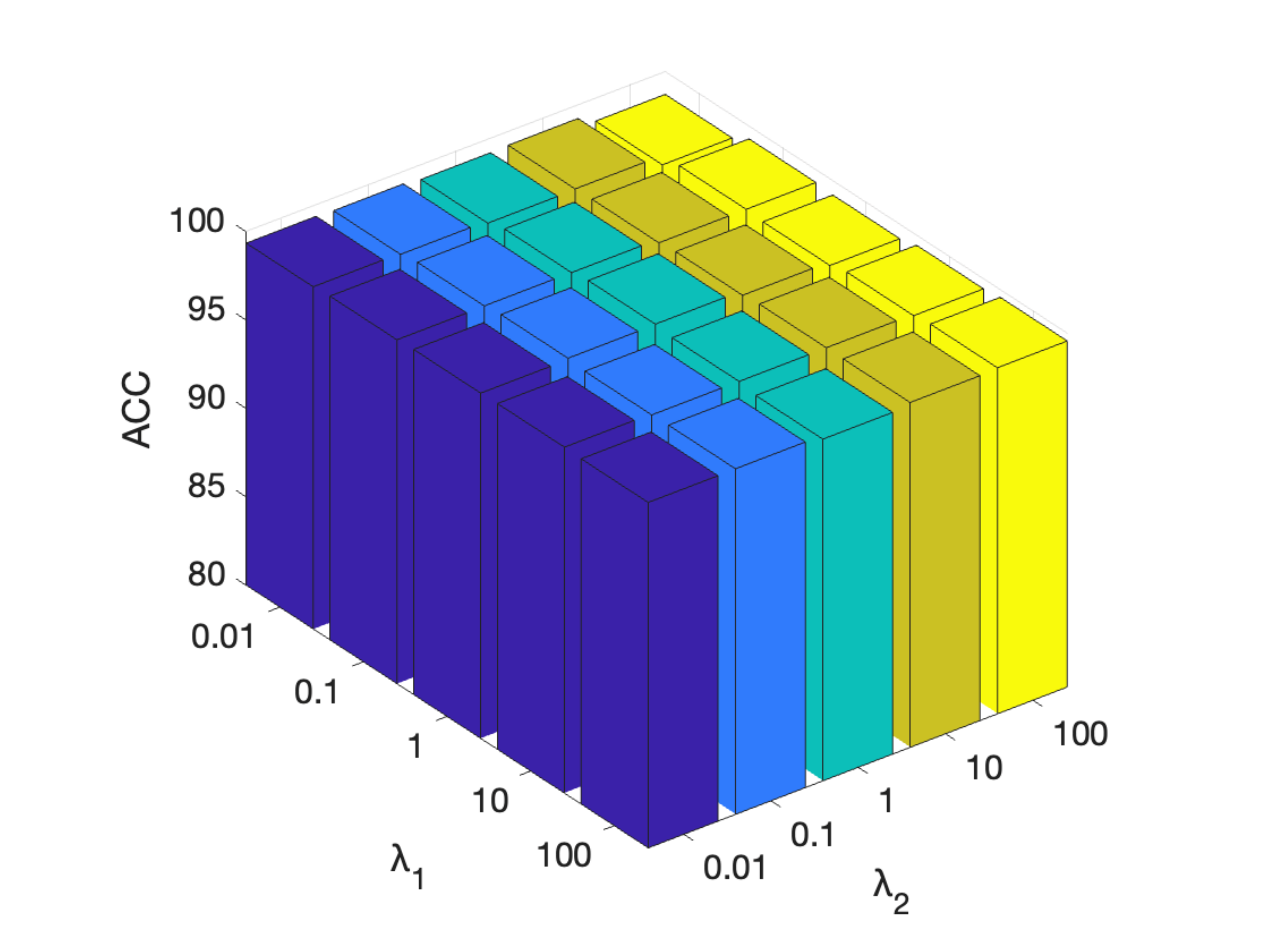}
        }
        \quad
        \subfigure[NMI in clustering task]{
        \includegraphics[width=0.45\linewidth]{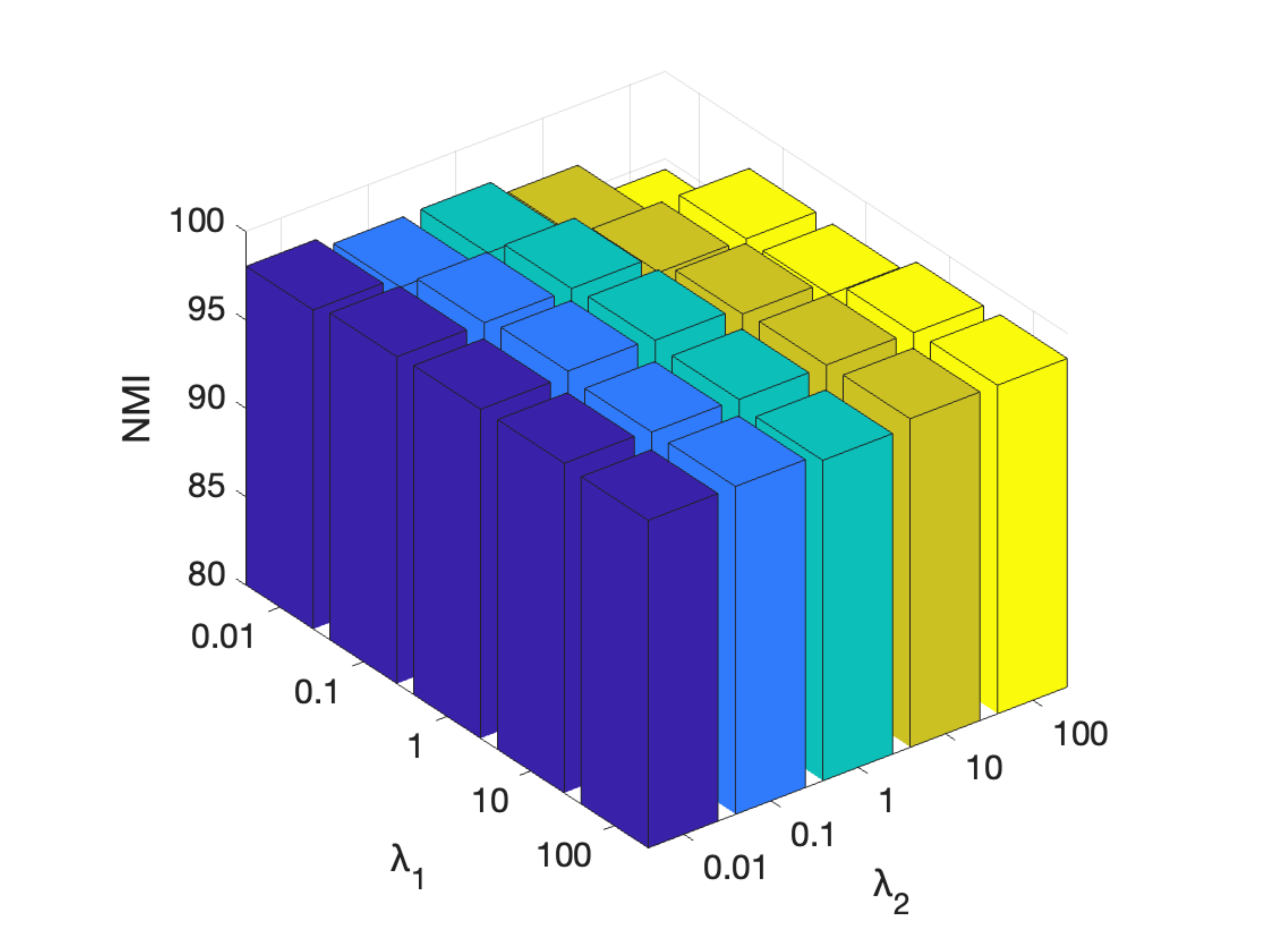}
        }
        \quad
        \subfigure[Sensitivity of model to parameter $\tau$ in classification task]{
        \includegraphics[width=0.45\linewidth]{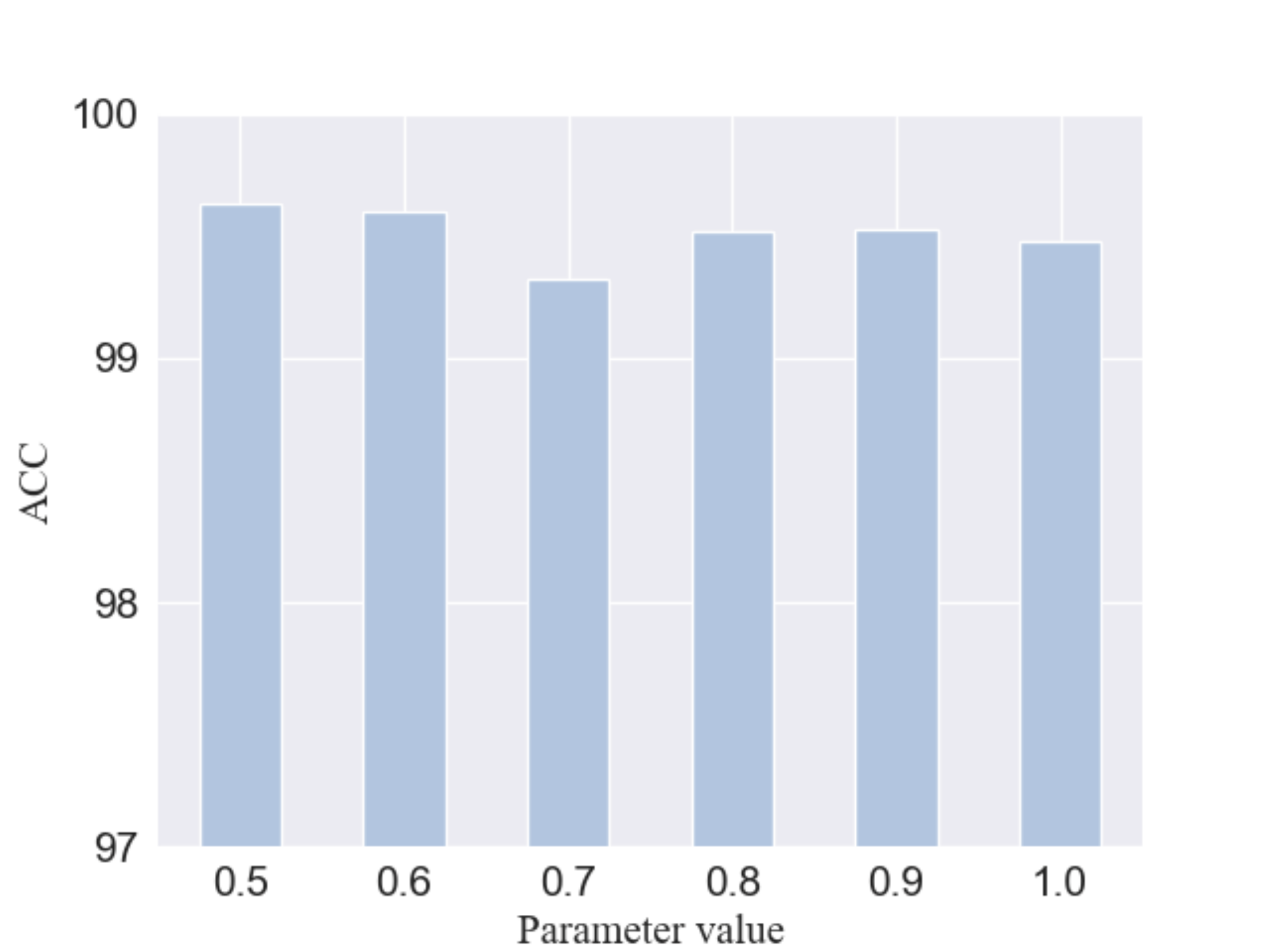}
        }
        \quad
        \subfigure[Sensitivity of model to parameter $\tau$ in clustering task]{
        \includegraphics[width=0.45\linewidth]{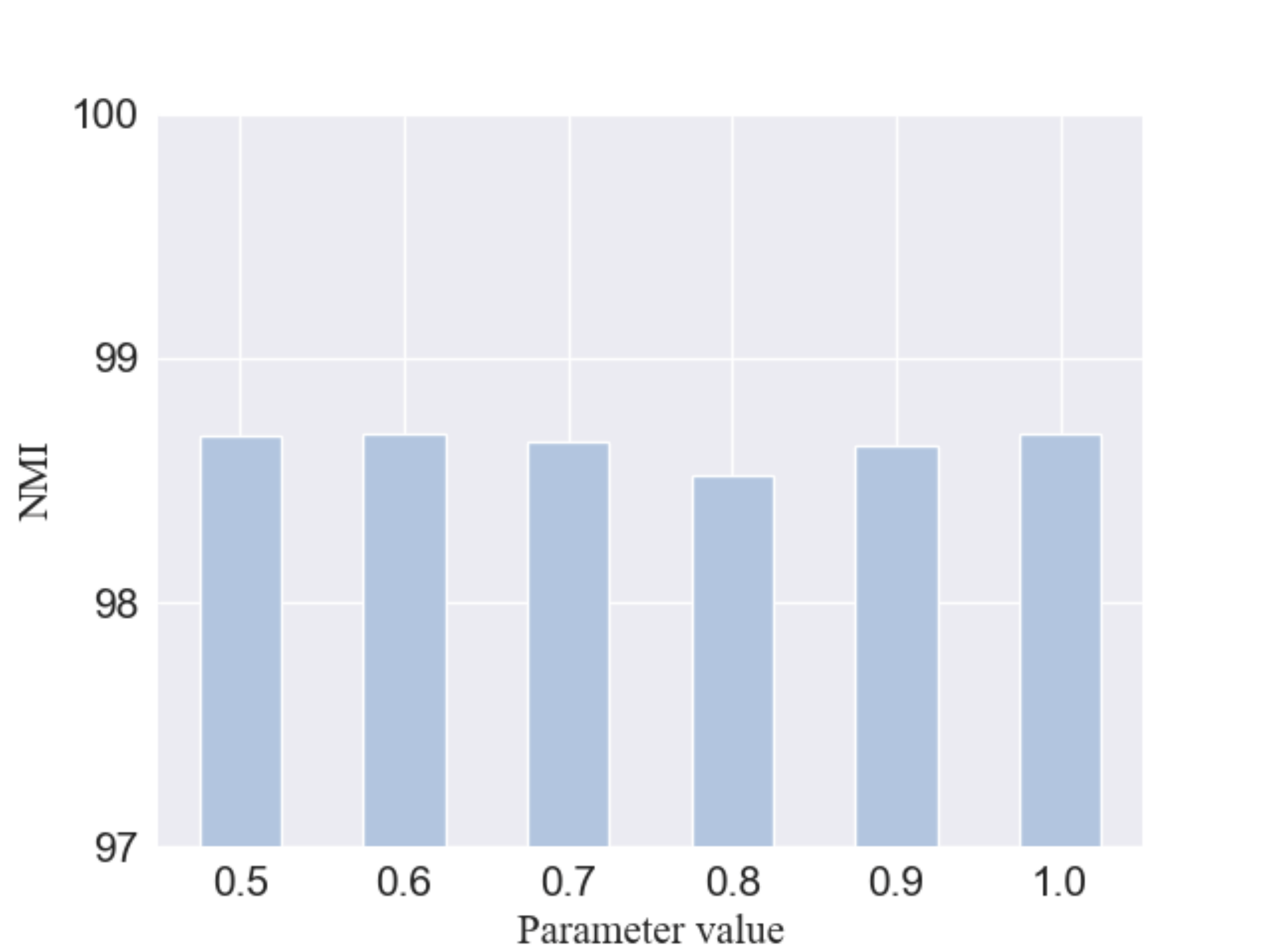}
        }
        \quad
        \caption{Parameter sensitivity analysis on MNIST-USPS dataset.}
	\label{fgg3}
\end{figure*}

\begin{table}[htbp]
\caption{Ablation study on Fashion dataset.
"\checkmark" indicates the used component, $Clu$ and $Cla$ denote the $Clu$stering task and the $Cla$ssification task respectively.}

\label{table5.}
\center
\begin{tabular}{r|cc|cc|c}
\toprule
\multirow{2}{*}{Dataset}    & \multirow{2}{*}{$\mathcal{L}_{Sem}$}          & \multirow{2}{*}{$\mathcal{L}_{Rec}$} &    \multicolumn{2}{c|}{$Clu$}    & $Cla$            \\ 
& & & NMI & Fscore & ACC\\
\midrule
 & \checkmark & \checkmark & \textbf{98.72} & \textbf{99.12} & \textbf{99.63} \\
\multicolumn{1}{r|}{Fashion} &                           & \checkmark & 81.69          & 76.19          & 93.57          \\
                                & \checkmark &                           & 96.74               & 97.34               & 98.49               \\ \bottomrule
\end{tabular}
\end{table}

\subsection{\bf Model Analysis}
\subsubsection{\bf Ablation Studies}
We conduct ablation studies here. It is clear that the module of unified feature representation learning is essential since UMRL aims to learn the unified multi-view representation. Consequently, we discuss the learning process of our method with and without $\mathcal{L}_{Sem}$ and $\mathcal{L}_{Rec}$. We take the experiments on the Fashion dataset for example. The clustering results (in metrics of NMI and Fscore) and the classification results (in the metric of ACC with \#Train/\#Test = 80\%/20\%) are reported in Table \ref{table5.}. According to the experimental results, we conclude: 1) Both the employment of $\mathcal{L}_{Sem}$ and $\mathcal{L}_{Rec}$ effectively improve the learned unified representation; 2) Compared with the employment of $\mathcal{L}_{Rec}$, the employment of $\mathcal{L}_{Sem}$ can significantly boost the performance of the learned unified representation.

In addition, we further discussed how much performance improvement of multi-view data compared with single-view data on downstream tasks and how different ways of integrating multi-view information affect the performance of downstream tasks.
Taking the clustering task as an example, we compare SCMRL and its different variants with the classical single-view clustering algorithm \cite{xie2016unsupervised} on BDGP and MNIST-USPS, and the results are shown in Table \ref{table_ab}.
In which, we use SCMRL\_concat to directly splice the low-dimensional representations $\{\mathbf{Z}^{(i)}\}_{i=1}^m$ of different views for clustering, while SCMRL\_average is to average $\{\mathbf{Z}^{(i)}\}_{i=1}^m$ of different views for clustering.

Convincingly, these aforementioned observations of ablation studies indicate that introducing excavation of multi-view semantic consistent information in the semantic space plays a vital role in our method for the learning process of multi-view unified feature representation.
Multi-view data has more descriptive information than single-view data, which is helpful for downstream tasks.
At the same time, in order to better integrate the data of multiple views, SCMRL's unique fusion method is more effective than simple splicing and averaging.

\begin{table*}[]
\centering  
\caption{Performance comparison between single view and multi-view.}  
\label{table_ab}
\begin{tabular}{ccccc}
\toprule
Dataset         & \multicolumn{2}{c}{MNIST-USPS}  & \multicolumn{2}{c}{BDGP}        \\ \hline
Metrics         & ACC            & NMI            & ACC            & NMI            \\ \hline
$k$-means (view 1) & 76.78          & 72.33          & 45.00          & 26.49          \\
$k$-means (view 2) & 58.10          & 52.13          & 57.04          & 45.96          \\
DEC (view 1)    & 73.10          & 71.49          & 46.28          & 29.96          \\
DEC (view 2)    & 56.12          & 61.08          & 94.78          & 86.62          \\
SCMRL\_concat   & 82.36          & 76.98          & 71.00          & 62.03          \\
SCMRL\_average  & 89.48          & 91.62          & 81.72          & 67.22          \\
SCMRL           & \textbf{99.56} & \textbf{98.72} & \textbf{98.00} & \textbf{94.69} \\ \bottomrule
\end{tabular}
\end{table*}

\subsubsection{\bf Parameter sensitivity analysis}
To explore the sensitivity of SCMRL to hyper-parameters, we take different values of $\lambda _{1}$ and $\lambda _{2}$ in Eq. (\ref{ep2}) on MNIST-USPS dataset and explore their influence of the clustering task (in the metric of NMI) and the classification task (in the metric of ACC).
In order to make the results more reliable, we run the clustering task and the classification task 10 times and 30 times respectively to average. The results are shown in Figure \ref{fgg3}(a) and Figure \ref{fgg3}(b), which indicates that our model is insensitive to $\lambda _{1}$ and $\lambda _{2}$. Actually, $\lambda _{1}$ and $\lambda _{2}$ are set to 1 for all datasets in Table \ref{table1.}-\ref{table4.}. 

As for the hyper-parameter $\tau$ in Eq. (\ref{ep11}), results of the MNIST-USPS datasets are shown in Figure \ref{fgg3}(c) and Figure \ref{fgg3}(d). It can also be found that SCMRL is also robust to $\tau$. Actually, we fix parameter $\tau =  0.5$ for all datasets in Table \ref{table1.}-\ref{table4.}. 

\begin{figure}[htbp]
\centering
\includegraphics[width=0.9\textwidth]{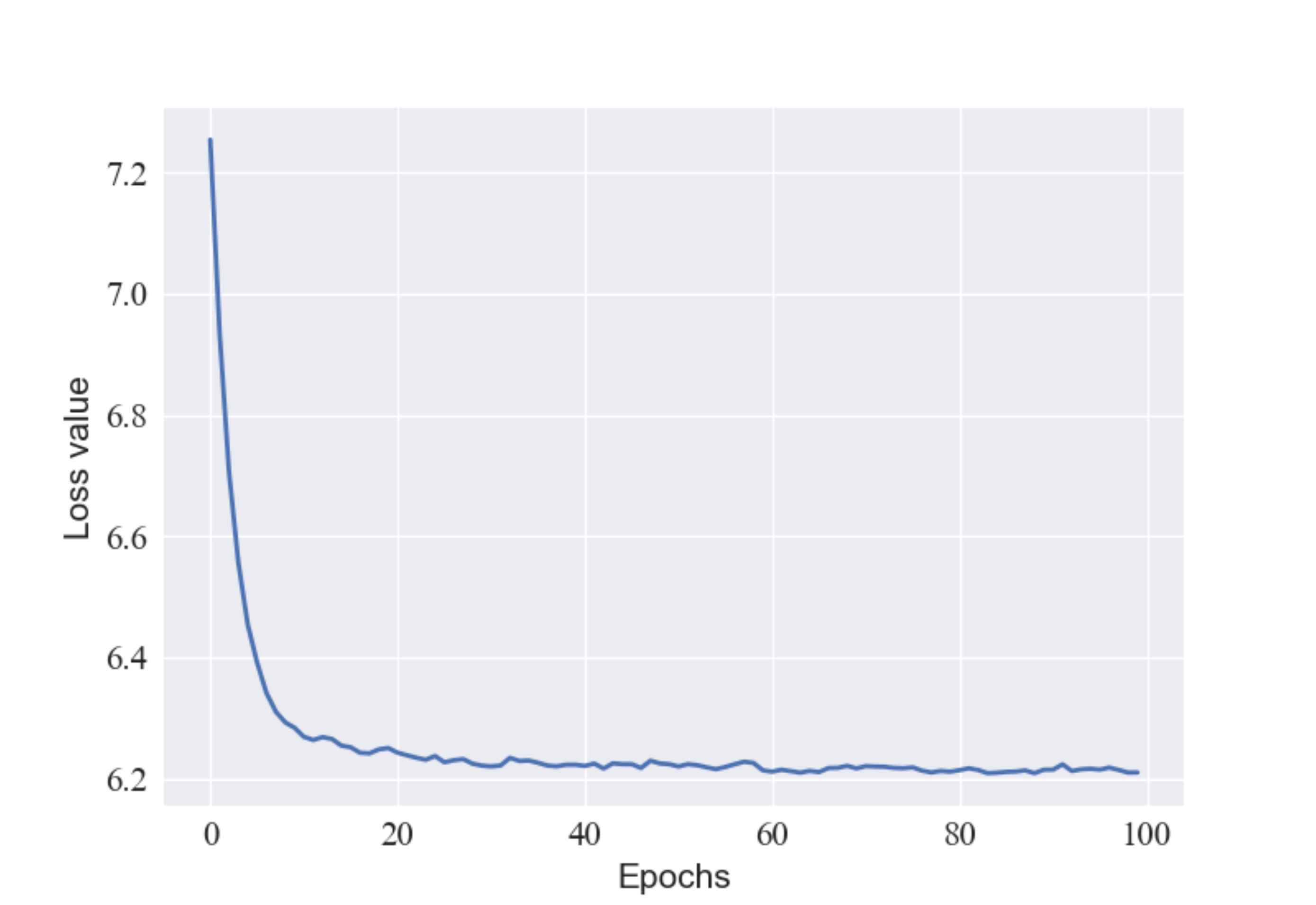} 
\caption{Convergence curve on MNIST-USPS dataset.} \label{fgg4}
\end{figure}

\subsubsection{\bf Convergence analysis}
To show the convergence properties of SCMRL, we take the experiment on MNIST-USPS dataset for example and display the convergence curve in Figure \ref{fgg4}. It can be observed that the loss value drops rapidly in the first 20 epochs and has promising convergence properties. For other datasets, the similar convergence properties can be achieved as well.

\section{Conclusion}
In this paper, we introduce a novel Semantically Consistent Multi-view Representation Learning (SCMRL), which effectively excavates and exploits the consistent information of multiple views in the semantic space to learn unified multi-view feature representation with promising structure.
By introducing the contrastive learning of semantic consistency, SCMRL incorporates the within-view reconstruction with the unified feature representation learning and explores the valuable consensus information in semantic space to guide the learning process.
Experimental results conducted on several benchmark datasets verify the effectiveness of SCMRL over other state-of-the-art methods.



\bibliographystyle{elsarticle-num} 
\bibliography{elsarticle-template-num}





\end{document}